\documentclass[graybox]{svmult}

\usepackage{amsmath}
\usepackage{amsfonts}
\usepackage{graphicx}
\usepackage{array}
\usepackage{caption}
\usepackage{microtype}
\usepackage{algorithm}
\usepackage{algpseudocode}
\usepackage{microtype}
\usepackage{breakcites}

\spnewtheorem*{notation}{Notation}{\itshape}{\rmfamily}

\usepackage{mathptmx}       
\usepackage{helvet}         
\usepackage{courier}        
\usepackage{type1cm}        
\usepackage{graphicx}        
\usepackage{multicol}        
\usepackage[bottom]{footmisc}
\usepackage[english]{babel}

\begin{document}

\title*{A Study of the Spatio-Temporal Correlations in Mobile Calls Networks}
\author{Romain Guigour\`es, Marc Boull\'e and Fabrice Rossi}
\institute{Guigour\`es, Boull\'e \at Orange Labs \at 2 av. Pierre Marzin \at 22300 Lannion - France \at \email{romain.guigoures@orange.com, marc.boulle@orange.com}
\and Guigour\`es, Rossi  \at SAMM EA 4543 - Universit\'e Paris 1 \at 90 rue de Tolbiac \at 75013 Paris - France \at \email{romain.guigoures@malix.univ-paris1.fr , fabrice.rossi@univ-paris1.fr}}
%
%
\maketitle

\abstract{
For the last few years, the amount of data has significantly increased in the companies. It is the reason why data analysis methods have to evolve to meet new demands. In this article, we introduce a practical analysis of a large database from a telecommunication operator. The problem is to segment a territory and characterize the retrieved areas owing to their inhabitant behavior in terms of mobile telephony. We have call detail records collected during five months in France. We propose a two stages analysis. The first one aims at grouping source antennas which originating calls are similarly distributed on target antennas and conversely for target antenna w.r.t. source antenna. A geographic projection of the data is used to display the results on a map of France. The second stage  discretizes the time into periods between which we note changes in distributions of calls emerging from the clusters of source antennas. This enables an analysis of temporal changes of inhabitants behavior in every area of the country.
}

\newpage
\section{Introduction}

The telecommunication operators interest in investigating the behavior of the customers using the call detail records has continuously grown in recent years. Several studies has been performed, some of them focusing on clustering antennas using the call flows \cite{Blondel2010}, \cite{Guigoures2011}. They highlight a strong correlation between the retrieved clusters and the covered territories characteristics like the spoken languages, the metropolitan areas in country-wide studies ; or the socio-economic profile of the neighborhoods (e.g. student, upper- or working-class) in local focuses. Such analysis are interesting for the telecommunication operators, particularly in developing countries where the needs in access to telecom services are becoming increasingly important while their usage are still unknown.

To go even further in the study of call detail records, clustering antennas from which the traffic similarly occurs over a studied time period could be investigated. A temporal analysis of the calls gives the means for understanding where excesses and lacks of traffic are located over the territory in function of the time period. Such a study provides information as well on the structure of the day, the week, the month or the year, as on the areas where the temporal phenomena are observed.

One major issue in the analysis of call detail records is the large amount of data. The data set we investigate in this article is a daily record of inter-antennas calls made in France from May 13, 2007 to October 13, 2007. The number of antennas throughout the French territory is $17{,}895$ between which $1{.}12$ billions calls have transited. The calls originating from (resp. terminating) outside the french mobile network are not included in the data. In Section~2, we introduce methods dealing with this kind of analysis and justify the choice of them for our problem. Then in Section~3, results on the spatial correlations are investigated while in Section~4 the temporal correlations are explored. Finally, the last section provides an assessment of the analysis results.

\section{Antenna Clustering based on Mobile Calls}
The first concern is the data representation. Indeed, a call is described by the source antenna, the destination antenna and the day it has been made. In a previous work \cite{Blondel2010}, an undirected graph is used to model a network of antennas linked by edges weighted by the calls frequency. In this paper, we choose to keep the matrix representation to exploit the natural direction of the calls.

\subsection{Related Works}

In \cite{Blondel2010}, the authors build a partition of the graph by
modularity maximization. This criterion \cite{Newman2006} measures the quality
of the segmentation of the graph into \textit{cliques} (or \textit{community}
in the graph-theory terminology) that are groups of strongly connected
vertices. The clusters of antennas obtained using this technique can be
efficiently retrieved by employing algorithms that exploit the sparsity of the
network \cite{Blondel2008} and thus modularity maximization is suitable for
problems with a large amount of data, like clustering antennas. Numerous other
graph clustering approaches have been proposed, see for instance
\cite{Schaeffer:COSREV2007,FortunatoSurveyGraphs2010} for surveys. However,
most of those approaches are based on some modularity or clustering
assumptions: in terms of our context, those assumptions mean that the calls
between antennas mostly occur within the groups of antennas. If it is the
case, using e.g. modularity maximization or other fast graph clustering
technique is very effective. If this is not the case, some patterns might be
missed and the actual structure of the graph not retrieved. In our problem, we
have no a priori knowledge that would justify a restriction to modular
patterns. In fact we have even reasons to believe that non symmetric and non
modular patterns might be present in the data: for instance, some antennas
might be associated to specific locations (universities, popular touristic
destinations) that lead to a significant amount of calls outside the area,
while other antennas might exhibit more localized call destinations.  We thus
must find an alternative approach that enables the discovery 
of any kinds of patterns. 

The concept of \emph{blockmodeling} originates in the pioneering works on
quantitative graph structure analysis conducted by sociologists in the 1950s
in the context of social network analysis \cite{Nad57}. To track the
underlying structure of the network, a matrix representation of a graph is
usually exploited, generally its adjacency matrix. Rows and columns represent
the source and destination antennas, and the values of the matrix indicate the
number of calls made between the antennas. Early sociological approaches
suggested to rearrange the rows and the columns in order to partition the
matrix in homogeneous blocks, a technique called \emph{blockmodeling}. Once
the blocks are extracted, a partition of the antennas of both source and
destination subsets can be deduced. This type of simultaneous grouping is
named \emph{co-clustering}. Notice that the only way to produce non symmetric
patterns (between source and target antennas) is to allow for two different
clusterings (one for the source antennas and one for the destination
antennas), thus leading to a co-clustering. Using this technique, we are able to track more
sophisticated patterns than approaches based on a single clustering whose
quality is judged by a density based measure such as the modularity. In fact
such approaches can be considered as looking for a diagonal blockmodel in
which off diagonal terms should be zeros. 

Numerous methods have been proposed to extract satisfactory clusters of vertices. Some of them \cite{Doreian2004} are based on the
optimization of criteria that favor partitions with homogeneous blocks, especially with pure zero-blocks as recommended in \cite{White1976}. More recent deterministic approaches have focused on optimizing criteria that quantify how well the co-clustering summarizes the input data \cite{Reichardt2007} (see e.g. \cite{Wasserman1994} for details on such criteria). Other approaches include \emph{stochastic} blockmodeling. In those generative models, a latent cluster indicator variable is associated to each
vertex. Conditionally to the latent variables, the probability of observing an
edge between two actors follows some standard distribution (a Bernoulli
distribution in the simplest case) whose parameters only depend on the pair of
clusters designated by the latent variables. In early approaches, the number
of clusters is chosen by the user \cite{Snijders2001}. More recent techniques
automatically determine the number of clusters using a Dirichlet Process
\cite{Kemp2006}. Finally, some recent approaches consider non-boolean latent
variables: cluster assignments are not strong and a vertex has an affiliation
degree to each cluster \cite{Airoldi2008}.

In addition to the diversity of structures that can be inferred from the network, co-clustering approaches are also able to deal with continuous variables \cite{Nadif2010},\cite{Boulle2012}. Blocks are extracted from the data that yields a discretization of the continuous variables. For a further analysis, we are able to track temporal patterns: the source antennas are still the rows in the data matrix while the columns now model the time.

In the case of an analysis of a call detail record, the technique we employ must have some properties:

\begin{itemize}
\item \textbf{Scalability}: with nearly $18{,}000$ antennas and $1{.}12$ billion calls, we cannot afford to use methods with a too high algorithmic complexity, that is often an issue with co-clustering/blockmodeling techniques.

\item \textbf{Genericity}: the processed data are either nominal or continuous. This point is really important in our study because we focus on nominal attributes (the antennas label) and continuous (the time).

\item \textbf{User-parametrization free}: data are complex and their underlying structure is a priori unknown, giving parametrization of the co-clustering scheme (e.g number of clusters, etc.) might be an issue for the user with such a data set. 

\item \textbf{Reliability} : the chosen approach must not yield spurious patterns, be resilient to noise and avoid overfitting. 

\item \textbf{Fineness and interpretability} : the approach must exploit all the relevant data information in order to extract fine patterns. In addition, exploratory analysis tools must allow users to work with the results effectively. \\

\end{itemize}

Given the large amount of data, the majority of the co-clustering approaches is not applicable to the problem of antennas clustering. Sampling the data set might be possible. However with $17{,}895$ antennas and $1{.}12$ billion calls, the average frequency of calls between two antennas is approximately $3{.}5$ and sampling the data would lead to a significant loss of information. Among the co-clustering approaches, we decide to use the MODL approach \cite{Boulle2010} \footnote{Software available on \url{www.khiops.com}} .

\subsection{Applying the MODL Approach}

Before detailing the chosen approach, the problem must be formalized. The data
set under study consists of \emph{calls} which are the statistical units. Each
call is described by three variables: the source antenna of the call, an
element of $V_S$ (see Table~\ref{tab:notationS}); the destination antenna of
the calls, an element of $V_D$; and the time at which the call started a real
number (the observed call times form the set $V_T\subset\mathbb{R}$). Notice
that while physical antennas are handling both incoming and outgoing calls,
those two roles are separated in our model: $V_S$ and $V_D$ are completely
distinct sets. This allows to build a directed model of the phone calls and
therefore limits information loss.

The Table~\ref{tab:notationS} lists the data features and the modelization
parameters we want to infer. The analysis we perform can be divided into two
steps. In the first one, we focus on the correlations between source and
destination antennas while in the second, we concentrate on studying the time
dimension of the calls. That is why, we introduce two distinct models:
one is spatial $\mathcal{M}_S$ and the other one is temporal
$\mathcal{M}_T$. In both case, the MODL approach infers the  parameters of the
model $\mathcal{M}_S$ (resp. $\mathcal{M}_T$) from the data $\mathcal{D}$. 

\begin{table}[!ht]%
\footnotesize
\begin{tabular}{|p{0.32\textwidth}|p{0.32\textwidth}|p{0.32\textwidth}|}
\hline
\textbf{$\mathcal{D}$ : Data} & \textbf{$\mathcal{M}_S$ : spatial co-clustering model}& \textbf{$\mathcal{M}_T$ : temporal co-clustering model}\\
\hline
 $V_S$: source antennas & $V_S^M$: partition of $V_S$ into  clusters of source antennas& $V_S^M$: partition of $V_S$ into  clusters of source antennas\\ & & \\
 $V_D$: destination antennas & $V_D^M$: partition of $V_D$ into  clusters of destination antennas&\\ & & \\
 $V_T$: time & & $V_T^M$: discretization of $V_T$ into time segments\\ & & \\
& $k_S$: number of clusters $V_S^M$& $k_S$: number of clusters of $V_S^M$\\ & & \\
& $k_D$: number of clusters of $V_D^M$&\\ & & \\
& & $k_T$: number of time segments of $V_T^M$\\ & & \\
& $k=k_Sk_D$: number of  biclusters& $k=k_Sk_T$: number of  biclusters\\ & & \\
\hline
 $n_S$: number of source antennas & $n^M_{i.}$: number of source  antennas in the $i^{th}$ cluster from the partition $V^M_S$& $n^M_{i.}$: number of source  antennas in the $i^{th}$ cluster from the partition $V^M_S$\\ & & \\
 $n_C$: number of destination  antennas & $n^M_{.j}$: number of destination  antennas in the $j^{th}$ cluster from the partition $V_D^M$&\\ & & \\
\hline
 $m$: total number of calls & &\\ & & \\
 $m_{i..}$: number of calls  originating from the source antenna $v_i$ & $m^M_{i..}$: number of calls  originating from the  $i^{th}$ cluster from the partition $V^M_S$& $m^M_{i..}$: number of calls originating from the $i^{th}$ cluster from the partition $V^M_S$\\ & & \\
 $m_{.j.}$: number of calls  terminating in the destination antenna $v_j$ & $m^M_{.j.}$: number of calls terminating in the $j^{e}$ cluster from the partition $V_D^M$ &\\ & & \\
& &  $m^M_{..t}$: number of calls made during the $t^{th}$ time segments  \\ & & \\
 $m_{ijt}$: number of calls made from the antenna $v_i$ to the antenna $v_j$ at time $v_t$& $m^M_{ij.}$: number of calls made from the $i^{th}$ cluster of source antennas to the $j^{th}$ cluster of destination antennas &  $m^M_{i.t}$: number of calls made from the $i^{th}$ cluster of source antennas during the $t^{th}$ time segment\\ & & \\
\hline
\end{tabular}
\caption{Notations.}
\label{tab:notationS}
\end{table}

In a first step, the model is based only on the antenna variables (source and
destination). The co-clustering approach is applied to the call detail record
to extract clusters of source antennas (in rows of the data matrix) and
destination antennas (in columns of the data matrix). The objective is to
group source antennas for which the calls are similarly distributed over the
destination antennas and conversely for target antenna w.r.t. source antenna.

In a second step, the model is based again on two variables: the source
antenna and the starting time of the call. As the time variable is continuous,
the clustering has been constrained to respect the time ordering. This
corresponds to a time quantization. The aim of the co-clustering in this case
is to simultaneously group antennas and discretize the studied time period
into segments during which the network is stationary. A higher order
co-clustering (e.g. a tri-clustering approach like in
\cite{guigouresICDM2012}) could be applied in order to keep the three original
variables. However, as will be become clear in Section \ref{sec:ST}, the
source/destination coupling is very strong in this data set, up to a point
where it hides the temporal patterns. By removing the destination variable,
one can hope finding temporal structures. 

MODL optimizes a criterion to find the co-clustering structure. The detailed formulation of the criterion as well as the optimization algorithms and the asymptotic properties are detailed in \cite{Boulle2011} for a co-clustering with nominal variables and in \cite{Boulle2012} for a co-clustering with heterogeneous variables, i.e nominal and continuous. The criterion is formulated following a MAP (Maximum a Posteriori) approach and is made up of a prior probability on the parameters of the co-clustering model and of the likelihood:

\begin{itemize}
	\item \textbf{The prior} : denoted $P(\mathcal{M}_S)$ (resp. $P(\mathcal{M}_T)$), it penalizes the model by specifying the a priori distribution of its parameters. It is hierarchically and uniformly built in order to be the most weakly informative \cite{Jaynes2003}. 
	
	\item \textbf{The likelihood} : Once the model parameters are specified, the likelihood $P(\mathcal{D}|\mathcal{M}_S)$ (resp. $P(\mathcal{D}|\mathcal{M}_T)$) is defined as the probability to observe the data given the parameters.\\

\end{itemize}

The product of the prior and the likelihood results in the posterior probability of the model. Its negative logarithm is the optimized criterion.

\begin{definition}
The spatial model $\mathcal{M}_S$, summarized representation of the data $\mathcal{D}$, is optimal if it minimizes the following criterion:
\begin{align}\label{eq:MODLS}
c(\mathcal{M}_S) &= -\log\left[ P(\mathcal{M}_S)\right] - \log\left[P(\mathcal{D}|\mathcal{M}_S) \right]   \notag\\
&= \log n_S + \log n_C + \log B(n_S,k_S) + \log B(n_C,k_D) + \log \binom{m+k-1}{k - 1}  \notag\\
&+ \sum_{c_i \in V_S^M} \log \binom{m^M_{i..} + n^M_{i.} -1}{n^M_{i.} -1} +
\sum_{c_j \in V_D^M} \log \binom{m^M_{.j.} + n^M_{.j} -1}{n^c_{.j} -1}  \\ 
&+ \log m! - \displaystyle\sum_{\substack{c_i \in V_S^M\\c_j \in V_D^M}} \log m^M_{ij.}! + \displaystyle\sum_{c_j \in V_D^M} \log m^M_{.j.}! - \displaystyle\sum_{v_j \in V_D} \log m_{.j.}!  \notag\\\notag
&+ \displaystyle\sum_{c_i \in V_S^M} \log m^M_{i..}! - \displaystyle\sum_{v_i \in V_S} \log m_{i..}!
\end{align}
\end{definition}

\begin{definition}
The temporal model $\mathcal{M}_T$, summarized representation of the data $\mathcal{D}$, is optimal if it minimizes the following criterion:
\begin{align}\label{eq:MODLT}
c(\mathcal{M}_T) &= -\log\left[ P(\mathcal{M}_T)\right] - \log\left[P(\mathcal{D}|\mathcal{M}_T) \right]  \notag\\
&= \log n_S + \log m + \log B(n_S,k_S) \notag\\
& + \log \binom{m+k-1}{k - 1} + \sum_{c_i \in V_S^M} \log \binom{m^M_{i..} + n^M_{i.} -1}{n^M_{i.} -1}   \\
&+ \log m! - \displaystyle\sum_{\substack{c_i \in V_S^M\\V_D^M \in V_T^M}} \log m^M_{i.t}! + \displaystyle\sum_{V_D^M \in V^M_T} \log m^M_{..t}!\notag\\\notag
& + \displaystyle\sum_{c_i \in V_S^M} \log m^M_{i..}! - \displaystyle\sum_{v_i \in V_S} \log m_{i..}! 
\end{align}
\end{definition}
$B(|V_S|,K_S)$ is the number of possible partitions of $V_S$ into $K_S$ potentially empty subsets. 

The two first lines of the equation~\ref{eq:MODLS} and the equation~\ref{eq:MODLT} are the prior terms while the two last lines are the likelihood terms. In an information-theoretic point of view, a negative logarithm of a probability amounts to a Shannon-Fano coding length \cite{Shannon1948}. Thus, the negative log of the prior probability $-\log(P(\mathcal{M}))$ is the description length of the model. As for the negative log likelihood $-\log(P(\mathcal{D}|\mathcal{M}))$, it is the description length of the data when modeled by the co-clustering. Minimizing the sum of these two terms therefore has a natural interpretation in terms of a crude MDL (minimum description length) principle \cite{Grunwald2007}.
The criterion $c(\mathcal{M})$ provides an exact analytical formula for the posterior probability of a model $\mathcal{M}$. That is why the design of sophisticated optimization algorithms is both necessary and meaningful. Such algorithms are described by \cite{Boulle2010}. 

The criterion is minimized using a greedy bottom-up merge heuristic. It starts from the finest model, i.e with one antenna per cluster and/or one timestamp per time segment. The merges of source and destination clusters or the merges of source clusters and time segments are evaluated and performed so that the criterion decreases. A post-optimization step improve this heuristic by making permutation between the clusters of antennas. This algorithm, that has a time complexity of $\mathcal{O}(m \sqrt{m} \log{m})$, is detailed in \cite{Boulle2010}.

\section{Analysis of the Spatial Correlations}
\label{sec:ST}

First, we focus on the analysis of the calls between source and destination antennas. We obtain $2{,}141$ clusters of source antennas and $2{,}107$ clusters of destination antennas. The average number of antennas per cluster is between $8$ and $9$, which is very fine. The challenge lies in exploiting the results. The number of clusters is too important for a countrywide analysis of the antennas grouping but is suitable for local studies. We thus propose analysis at different geographic scales.

\subsection{A Countrywide Analysis}

First, we propose a countrywide analysis of the results and a projection on a map of France. The finest results do not provide a summarized enough view of the co-clustering structure for such an analysis. That is why, we process an agglomerative hierarchical clustering of the clusters to reduce their number. Clusters are merged so that the criterion is the least decreased in order to obtain the most probable co-clustering model for a given number of clusters. This post-treatment allows a simplification of the model while handling its quality loss. In order to quantify this loss, we introduce a quality measure we call \textit{informativity rate}.

\begin{definition}
The null model $\mathcal{M}_S^{\emptyset}$ is the parametrization of the model, such that there is one single cluster of source and destination vertices or one single cluster of source and one time segment. The null model is the best representation of a data set with no underlying structure. Given the best model $\mathcal{M}_S^*$ obtained by optimizing the criterion defined in Definitions 1 and 2, the informativity rate of a model $\mathcal{M}_S$ is:

\[
\tau(\mathcal{M}_S)=\dfrac{c(\mathcal{M}_S)-c(\mathcal{M}_S^{\emptyset})}{c(\mathcal{M}_S^*)-c(\mathcal{M}_S^{\emptyset})}
\]

\end{definition}
By definition, $\tau(\mathcal{M}_S) \leq 1$ ; note that $\tau(\mathcal{M}_S) < 0$ is possible when $\mathcal{M}_S$ is an irrelevant modelization of the data $\mathcal{D}$ (e.g. $\mathcal{M}_S \neq \mathcal{M}_S^{\emptyset}$ when $\mathcal{D}$ are random data).

\begin{figure}[!ht]
	\centering
		\includegraphics[width=0.9\textwidth]{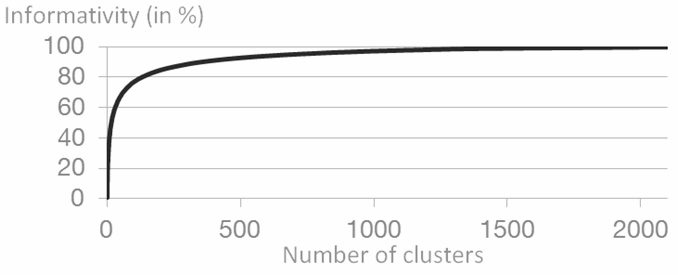}
	\caption{Informativity rate function of the number of clusters}
	\label{fig:pareto}
\end{figure}

The informativity rate allows the construction of a curve of the informativity of the model in function of the number of clusters. This aims at helping the user in finding a good trade off between a simple and an informative co-clustering.

\begin{figure}[!ht]
	\centering
		\includegraphics[width=0.9\textwidth]{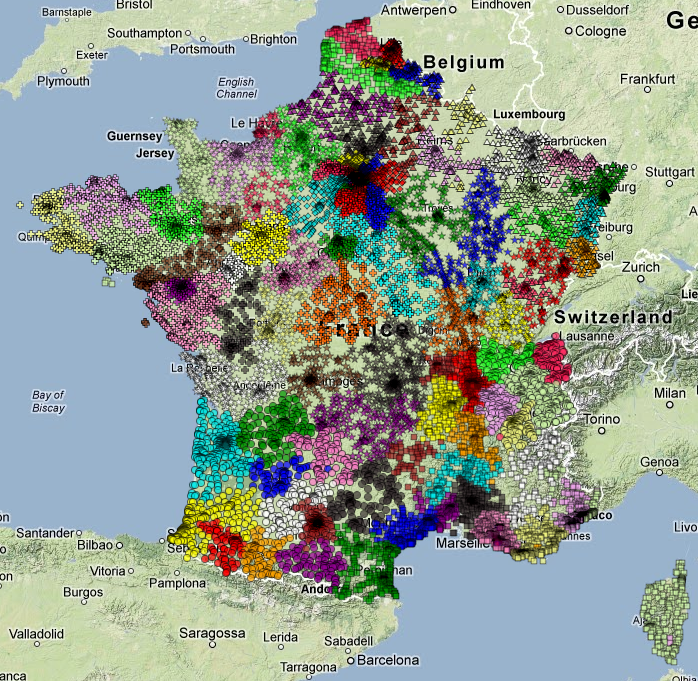}
	\caption{Projection of the clusters of source antennas on a map of France. There is one color and shape per cluster.}
	\label{fig:France2}
\end{figure}

The Figure~\ref{fig:pareto} shows that the first merges have a weak impact on the model informativity. Hence, the number of clusters of both source and destination antennas can be significantly reduce from more than $2{,}000$ to $85$ while keeping $75\%$ of the model informativity. This simplified model is used for the countrywide study: it is simple enough to be interpreted and informative enough to make a reliable analysis. Results are displayed in Figure~\ref{fig:France2}.

The correlation between the clusters of antennas and their locations is strong despite the antennas positions are not taken in account in the optimized criterion. We can thus deduce that the people living in the same areas use to call the same destination and vice-versa. The map of the Figure \ref{fig:France2} shows that the french territory can be segmented into several geographic areas that do not necessarily correspond to the regional administrative boundaries.

\subsection{A local analysis}

\begin{figure}[!ht]
	\centering
		\includegraphics[width=0.8\textwidth]{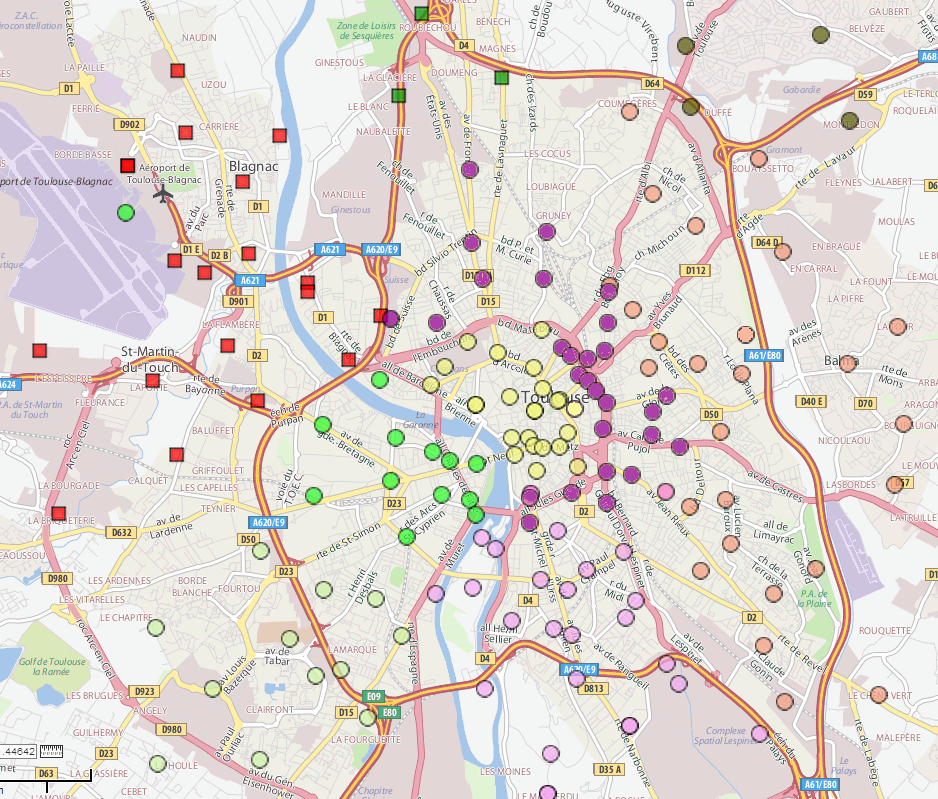}
	\caption{Projection of the clusters of source antennas on a map of Toulouse, there is one shape and color per cluster.}
	\label{fig:tlse}
	\end{figure}

In a second step, we focus on a local analysis. To that end, we exploit the finest model ($\mathcal{M}^*_S$) and only consider a subset of antennas corresponding to the area of a french city. The antennas in Toulouse are segmented into seven clusters, displayed in Figure~\ref{fig:tlse}. The first cluster groups the antennas of the inner city (pale yellow circles), another cluster (neon green circles) groups antennas located in the west bank of the Garonne river, that corresponds to a largely residential area. The cluster of antennas pictured as pale pink circles takes place over the University of Toulouse campus and a disadvantaged neighborhood. As for the cluster grouping antennas displayed using pale green circles, it covers an area with the same characteristics than the previous one. The orange circles are located in the residential periphery of the city with different socio-economic profiles: upper-class toward South and working-class toward North. Finally the red squares are antennas located in the industrial areas adjacent to the international airport.

In order to understand why antennas have been grouped together, we focus now on the distribution of calls originating from the clusters. To that end, we study the contribution to the mutual information of each couple of source/destination stations.

\begin{definition} This measure quantifies the dependence of two variables, here the partitions of the source and destination antennas. Let us denote it $MI(V_S^M,V_D^M)$, defined as follows \cite{cover}:

\begin{eqnarray}
MI(V_S^M,V^S_D) &=& \displaystyle\sum_{c_i^S \in V_S^M} \sum_{c_j^D \in V_D^M} p(c_i^S,c_j^D) \log \dfrac{p(c_i^S,c_j^D)}{p(c_i^S)p(c_j^D)} 
\label{eq:mi}
\end{eqnarray}
\end{definition}

Mutual information is necessarily positive and its normalized version is commonly used as a quality measure in the coclustering problems \cite{Strehl2003}. Here, we only focus on the involvement to mutual information of a couple of source/destination clusters stations. This value can be positive or negative according to whether the observed joint probability of journeys $p(c_i^S,c_j^D)$ is above or below the expected probability $p(c_i^S)p(c_j^D)$ in case of independence. Displaying such a measure allows to quantify whether there is a lack or an excess of calls between two groups of antennas in comparison to the expected traffic.

\begin{figure}[!ht]
	\centering
		\includegraphics[width=0.8\textwidth]{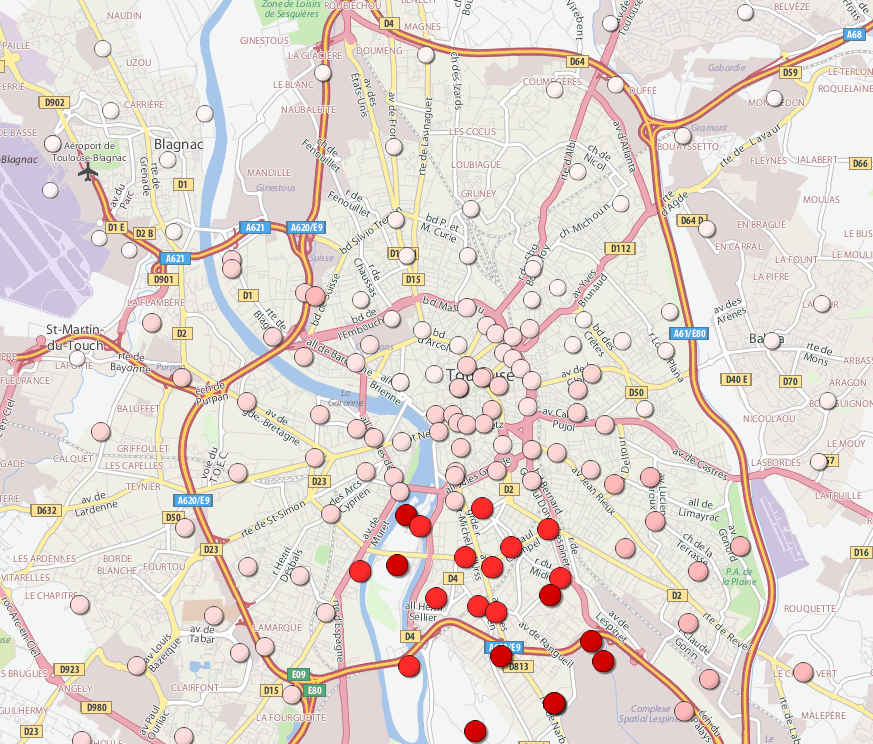}
	\caption{Contribution to the Mutual Information between the cluster of the university campus and the clusters of the plotted antennas.}
	\label{fig:tlseMI}
\end{figure}

This is illustrated in the Figure~\ref{fig:tlseMI}. We focus on the traffic of calls originating from the pale pink cluster of the Figure~\ref{fig:tlse}. Antennas that are pictures as red circles are the ones to which an excess of traffic from the studied cluster is observed ($p(V_S^M,V_D^M)>p(V_S^M)p(V_D^M)$) while the antennas corresponding to the white circles are the ones to which the traffic is null or expected ($p(V_S^M,V_D^M) \approx 0 \mbox{ or } p(V_S^M,V_D^M) \approx p(V_S^M)p(V_D^M) $). For this cluster of source antennas, there are no antennas to which we observe a significant lack of traffic. If any, their location would have been identified by a blue circle on the map. Note that the colors in the map represent the contribution to the mutual information, not the frequency of calls which logarithm is proportional to the diameter of the circles. Hence, we observe that the excess of traffic mainly occurs within the cluster we focus on, and slightly to the rest of the city. 
	
\section{Spatio-Temporal Analysis}

\begin{figure}[!ht]
\centering
		\includegraphics[width=0.8\textwidth]{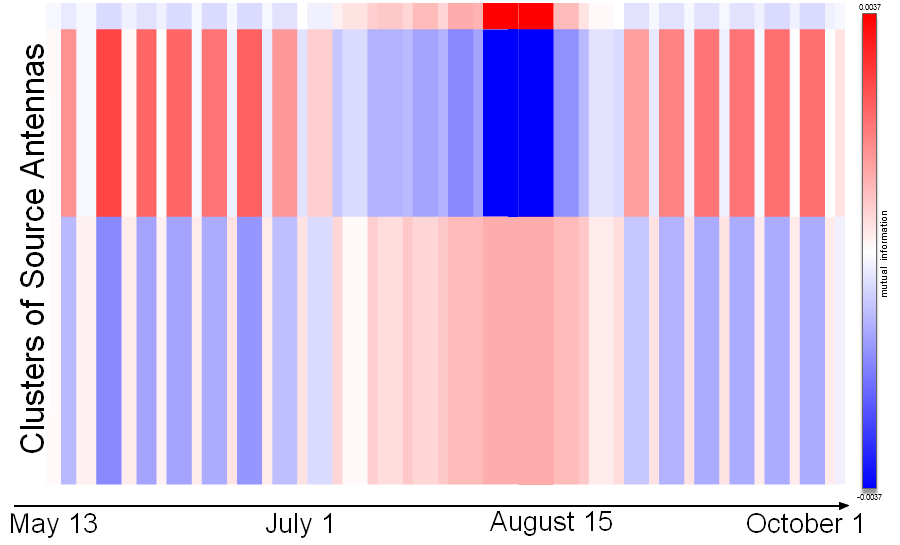}
	\caption{Calendar of the excesses and lacks of calls from three clusters and $42$ time segments.}
	\label{fig:calendar}
\end{figure}

In this second study, we propose to process a co-clustering on the source antennas and the time. In this study, we could have envisaged to apply a tri-clustering approach like in \cite{guigouresICDM2012}. However, the previous analysis showed us a strong correlation between the partitions of source and destination antennas. Hence, we consider that both source and destination antennas bring the same information and we consequently use only one of them. The data are call detail records with $17{,}895$ source antennas and $1{.}12$ billion calls made over five month. The timestamps are the dates. The antennas grouping is different from the one we obtained in the Section~\ref{sec:ST}. Here the antennas are grouped if the emerging calls are similarly distributed over the days. We obtain $6{,}129$ clusters of source antennas and $117$ time segments. Contrary to the source/destination antennas analysis, there is no correlations between the clusters of antennas and their locations: they are scattered over the entire french territory. As a consequence, a projection of the clusters on a map of France would not be interpretable, even for a reduced number of clusters. In order to investigate the phenomena that lead to such a result, we also study the contribution to the mutual information between the clusters of antennas and the time segments. To visualize this measure, we have simplified the co-clustering model in the same way as for the previous study and we have plotted a calendar of the excesses and lacks of traffic in Figure~\ref{fig:calendar}.

From May 13 to July 5 and from September 1 to October 13, the discretization is periodic, highlighting the working days and the week-ends. During the working days at both these periods, the calls originate in excess from the antennas of the middle cluster and in deficit from the bottom cluster. The contrast between the clusters is reducing as summer approaches. As for the weekends, the inverse phenomenon is observed, but not to the same degree. These phenomena can be explained by the agglomeration of the economic activity on concentrated geographical areas, usually urban. Note that there is always a lack of calls originating from the top cluster at this period.

\begin{figure}[!ht]
	\centering
		\includegraphics[width=0.8\textwidth]{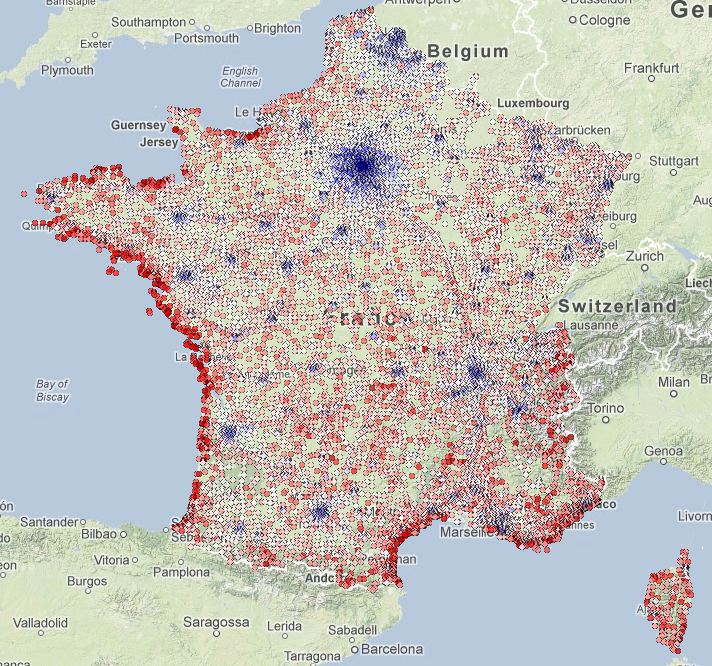}
	\caption{Contribution to the mutual information between the clusters of source antennas and the time segment from August 5 to August 15. In red, the antennas from which there is an excess of calls, in blue a deficit and in white the expected number of calls.}
	\label{fig:FranceMI}
\end{figure}

During the summer vacations, the periodicity working days/weekends is not observed anymore. The calls originating from the middle cluster are now in deficit while the ones made from the top cluster are significantly in excess compared to the usual traffic of the areas covered by the antennas and the traffic in the time segments. It is during this period that the contrast is the sharpest. That is the reason why, we focus on the segment from August 5 to August 15 and draw a map where the antennas are displayed and colored in function of the excess or deficit of calls made during this segment (see Figure \ref{fig:FranceMI}).

During the summer vacations, there is a significant excess of calls made from the Atlantic and Mediterranean coasts. This means that during this period, the number of calls originating from these areas are more important than usually. These areas can truly be regarded as seasonal regions since they are characterized by an unbalanced distribution of calls over the year. Actually the population of the areas covered by this cluster have also an unbalanced population over the year: during the summer vacations, the population can be multiplied by more than ten, this has a direct impact on the calls distribution.

Conversely, the cities are colored in blue on the map of the Figure~\ref{fig:FranceMI}. This can be explained by the diminution of the economic activity due to the vacations. It is thus assumed that the populations move from the urban centers to the holidays resorts located on the seashore. We note however that the color only indicates the contribution to the mutual information, not the frequency: during this period, the antenna from which the most calls have been made is colored in blue and located in Paris.

\section{Conclusion}

In this article, we have proposed an analysis of a five month call detail record between $17{,}895$ mobile phone antennas spread throughout the French territory. That represents a total of $1{.}12$ billion calls. After having listed similar studies and introduced methods suitable for such analysis, we have discussed on the choices that conducted us to use the MODL approach. Two different types of analysis have been conducted while using one single approach, being generic and scalable enough to thoroughly investigate the data. 

In a first study, the antennas have been grouped together if the calls originating from (resp. terminating to) them are distributed on the same groups of antennas. An analysis of a projection of the clusters on a map reveals a strong correlation between the geographic position of the antennas and the clusters they belong to, at the national or local levels. In a second study, we have lead a study in which the time is taken into account. Despite, the antennas belonging to a same cluster are not located in a well-defined area anymore, they cover nevertheless areas with common features: urban, rural or touristic. As for the time segmentation, this highlights different behaviors in terms of mobile phone usage during the summer vacations and the working periods, during which we observe a periodicity between the working days and the weekends. For example in August, there is an excess of calls in the touristic areas while there is a deficit of calls in the urban areas, where most economic activity is concentrated.

In future works, it might be interesting to lead a study in which several time features are embedded in order to characterize the behavior in terms of mobile phone usage, in function of the date, the day of the week and the time of the day. 

 \bibliographystyle{apalike}
 \bibliography{biblio} 
 \end{document}